\documentclass[11pt]{article}
\PassOptionsToPackage{table}{xcolor}
\usepackage[preprint]{acl}
\usepackage{enumitem}
\usepackage{times}
\usepackage{latexsym}
\usepackage{booktabs}
\usepackage{graphicx}
\usepackage{multirow}
\usepackage{colortbl}
\usepackage{float}
\usepackage[most]{tcolorbox}
\usepackage{xcolor}

\definecolor{groupblue}{RGB}{225,245,255}

\usepackage[T1]{fontenc}

\usepackage[utf8]{inputenc}

\usepackage{microtype}

\usepackage{inconsolata}
\usepackage{amsmath}

\usepackage{graphicx}

\title{See More, Think Deeper: Query-Expanded Visual Evidence and Answer-Clue Guided Reflection for Long Video Understanding}

  \author{
    Shuning Wang$^{*1}$, Zhiheng Wu$^{*\S 1}$, YiNuo Lu$^{*2}$,
    Naiming Liu$^{2}$, Chen Jia$^{1}$, Bowen Liu$^{3}$, \\
    \textbf{Shuo Nie$^{2}$, Weijie Zhu$^{1}$, Yumeng Zhang$^{\dagger 1}$} \\
    \\
    $^{1}$Baidu Inc.,
    $^{2}$Harbin Institute of Technology, \\
    $^{3}$Hong Kong University of Science and Technology
     \\
    \texttt{\small $^{*}$ Equal contribution. $^{\S}$ Project leader. $^{\dagger}$ Corresponding author.}
  }

\begin{document}
\maketitle
\begin{abstract}
Recent advances in Video Large Language Models (Video-LLMs) have enabled performance on long-video understanding tasks. However, existing methods still face two key limitations: evidence acquisition often relies on a single search intent, and answer generation lacks an effective visual feedback mechanism. To address these limitations, we propose \textbf{CoVER}, a Comprehensive Visual Evidence and Reflection framework for long-video understanding. CoVER enables Video-LLMs to \textbf{See More} by dynamically gathering query-expanded visual evidence, and \textbf{Think Deeper} by verifying draft answers with effective answer-specific visual feedback. Together, these mechanisms shift long-video understanding from answer-centric generation to evidence-centric and visually verifiable reasoning. Experimental results show that CoVER-7B substantially outperforms models with the same parameter scale and even surpasses state-of-the-art closed-source models on certain metrics.
\end{abstract}

\section{Introduction}
\begin{figure}[t]
    \centering
    \includegraphics[width=\columnwidth]{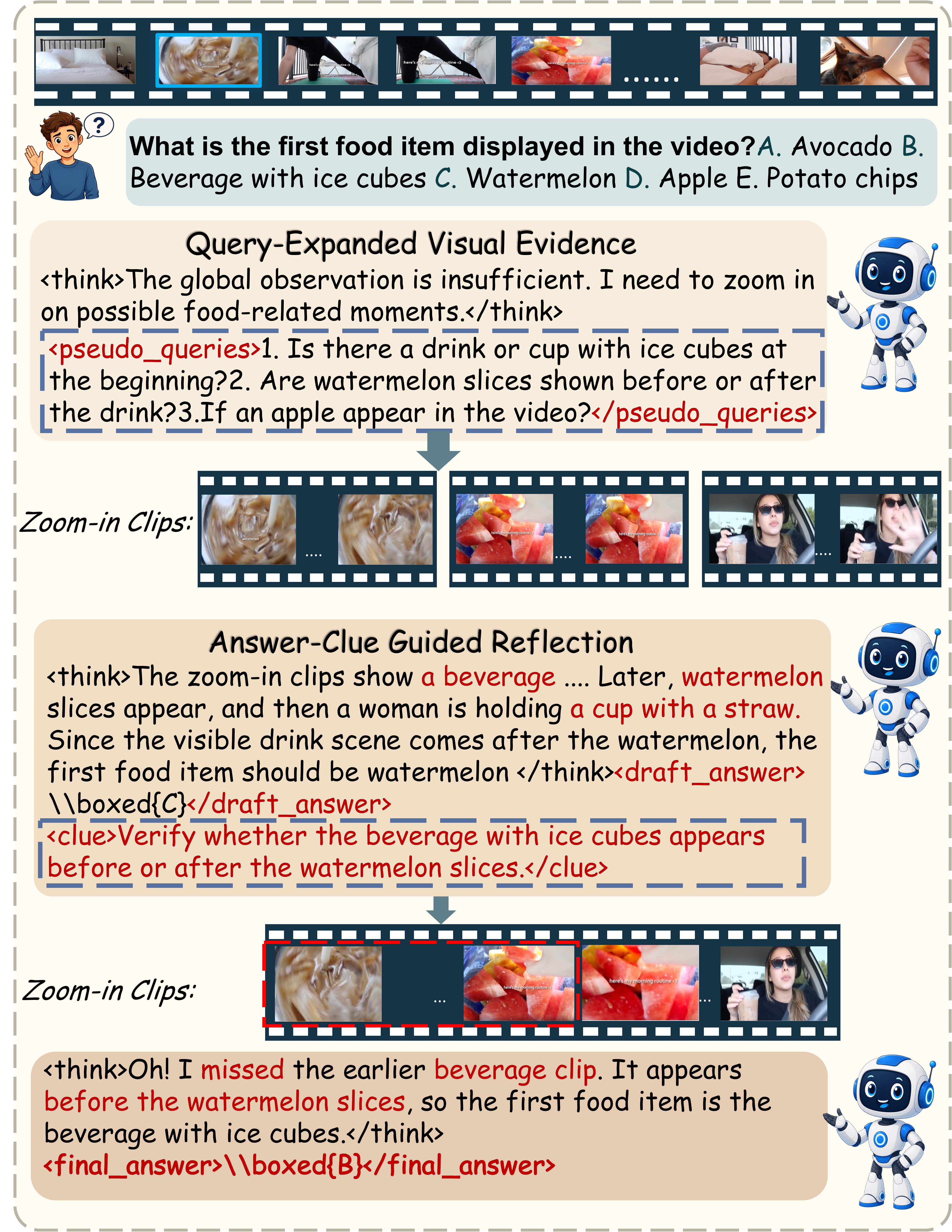}
    \caption{
\textbf{Illustration of CoVER reasoning process.}
CoVER retrieves task-relevant zoom-in evidence with pseudo-queries, then uses an answer-specific clue to verify and refine the draft answer with visual evidence.
}
    \label{fig:intro}
\end{figure}

Recent advances in Video Large Language Models (Video-LLMs) have greatly improved video question answering, event understanding, and long-video reasoning~\cite{fu2025love, lin2024video,ouyang2025conan,li2024llama}. These models can now interpret complex video content and respond in natural language~\cite{liu2025videomind}. However, long-video understanding is not merely about seeing more frames. It requires identifying sparse but crucial evidence across extended temporal contexts. Key evidence may appear only in a few segments, local regions, or brief moments. Thus, models must not only capture global context, but also locate fine-grained visual evidence during reasoning. Recent studies have therefore shifted from coarse-grained video understanding to fine-grained temporal or spatio-temporal grounding. For example, VideoChat-R1.5 uses Iterative Perception to progressively focus on high-confidence spatio-temporal regions~\cite{yan2026videochat}, while LongVT first skims the video globally and then inspects relevant clips with a video-cropping tool~\cite{yang2025longvt}.

Although these methods improve evidence localization, existing VideoQA~\cite{zhang2025thinking, liu2025longvideoagent} still face a fundamental challenge: a correct answer may rely on incorrect or incomplete evidence. Prior work on visually grounded VideoQA shows that high answer accuracy does not necessarily imply accurate evidence grounding~\cite{ouyang2025conan, wang2026think}. VideoZeroBench~\cite{meng2026videozerobench} further highlights this gap, showing that model performance drops sharply when both answer correctness and spatio-temporal evidence correctness are required. This suggests that current video reasoning often relies on linguistic priors, salient partial cues, or incomplete context, rather than complete and verifiable visual evidence.

One reason is incomplete evidence acquisition. Existing methods~\cite{wang2025vr,fu2025love} often retrieve evidence using only the original question or a single search intent. This may be sufficient for simple questions, but complex video reasoning usually requires multiple cues, such as object attributes, human actions, state changes, spatial relations, and temporal order. A single coarse-grained query may capture only salient cues while missing other critical details. To address this, we introduce query-expanded evidence acquisition. Given the original question, we generate diverse evidence-seeking queries that decompose the reasoning demand into observable and retrievable visual cues. In this paper, we refer to these model-generated evidence-seeking queries as pseudo-queries. This helps the model collect more comprehensive evidence before reasoning.

Another limitation is the lack of visual verification after answer generation. Existing Video-LLMs typically reason over available visual or textual context and directly produce an answer~\cite{wu2026chain, yang2026videodetective}. Once the answer is generated, the process stops. The model rarely checks whether the answer is supported by visual evidence, nor does it revise the answer when new evidence is found. To address this, we introduce answer-clue-guided visual reflection. After producing an initial answer, the model derives verifiable visual clues from the answer and retrieves related evidence from the video. The visual feedback is then used to re-evaluate and revise the answer when necessary.

Based on these insights, we propose \textbf{CoVER}, a \textbf{Co}mprehensive \textbf{V}isual \textbf{E}vidence and \textbf{R}eflection framework for long-video understanding. CoVER is equipped with a text-aware video evidence tool that retrieves relevant video segments and provides magnified visual information for reasoning. As shown in Fig.~\ref{fig:intro}, it contains two modules: (1) query-expanded visual evidence acquisition, which generates diverse evidence-seeking queries to help the model \textbf{See More}; (2) answer-clue-guided visual reflection, which retrieves evidence based on clues derived from the initial answer and uses visual feedback to \textbf{Think Deeper}.  Together, these modules shift long-video understanding from answer-centric generation to evidence-centric and visually verifiable reasoning. Experiments show that CoVER improves long-video understanding performance. Compared with Qwen2.5-VL-7B, CoVER-7B achieves gains of +3.9\% on MLVU and +4.6\% on LVBench.

\noindent
Our contributions are summarized as follows:
\begin{itemize}[leftmargin=*, itemsep=1pt, topsep=2pt, parsep=0pt, partopsep=0pt]
    \item We are the first to introduce visual-feedback reflection into Video-LLM reasoning, transforming open-loop inference into a closed-loop paradigm.
    \item We propose CoVER, a framework that combines query-expanded visual evidence acquisition with answer-clue guided reflection for evidence-centric reasoning.
    \item Experiments on long-video benchmarks show that CoVER brings substantial gains over the baseline.
\end{itemize}

\section{Related Work}
\textbf{Long Video Understanding.}
Large Video Language Models (LVLMs) have advanced and demonstrated strong capability across video understanding tasks~\cite{bai2025qwen3, zhang2410video, li2025videochat,li2024llava,lin2024video}. Existing approaches commonly adopt uniform frame sampling~\cite{maaz2024video}, where frames are sampled at fixed intervals and concatenated as model input. However, in long-video settings, this strategy leads to a rapid increase in visual tokens. To improve efficiency, prior studies explore token compression, keyframe selection, long-context modeling, and agent-based multi-stage processing~\cite{tang2025adaptive, wang2025dynamic,tang2026tspo,wang2026think}. Unlike these approaches, CoVER uses query expansion to adaptively retrieve task-relevant evidence and verifies draft predictions with answer-specific visual clues, enabling evidence-grounded long-video reasoning.

\noindent
\textbf{Chain-of-Thought for Multimodal Reasoning.}
Recent reasoning models show that long Chain-of-Thought (CoT) can improve complex reasoning by decomposing problems into intermediate steps~\cite{li2025towards,wei2023chainofthoughtpromptingelicitsreasoning,wu2026see}. 
For video understanding, prior studies further incorporate visual information into reasoning, enabling operations such as zoom-in, grounding, and evidence selection~\cite{zhang2025vitcot,jiang2025vlm,jian2025look}. 
Recent methods, including VITAL~\cite{zhang2025thinking}, LOVE-R1~\cite{fu2025love}, REVISOR~\cite{li2025revisor}, and Conan~\cite{ouyang2025conan}, explore dynamic video reasoning and multimodal reflection with tool use or multi-scale visual evidence. 
Other evidence-centric approaches study search-guided grounding, semantic-visual evidence consensus, and explicit spatio-temporal evidence for long-video answers~\cite{wu2026chain,sheng2025sevices,wei2026seeing}. 
Unlike these methods, CoVER unifies question-side evidence acquisition and answer-side verification in a closed-loop process, using pseudo-queries to expand visual evidence and answer-specific clues to refine draft predictions.

\noindent
\textbf{RL-based Reasoning in VLMs.} Recent progress in reinforcement learning (RL) and preference optimization has become an effective paradigm for improving reasoning in Vision-Language Models (VLMs)~\cite{guo2025deepseek,yu2025dapo,rafailov2023direct}. This trend has also been extended to video understanding, where existing efforts can be broadly grouped into three directions: optimizing temporal sampling for long-video understanding, such as TSPO~\cite{tang2026tspo}; improving temporally grounded and consistent reasoning, as in Video-R2~\cite{maaz2025video}; and enhancing dynamic multi-step video reasoning, as explored by LOVE-R1~\cite{fu2025love} and Video-Thinker~\cite{wang2025video}. Unlike these methods, CoVER focuses on evidence completeness and answer verification: it uses query expansion to retrieve task-relevant visual evidence and answer-specific clues to verify and refine draft predictions.

\begin{figure*}[t]
    \centering
    \includegraphics[width=0.9\textwidth]{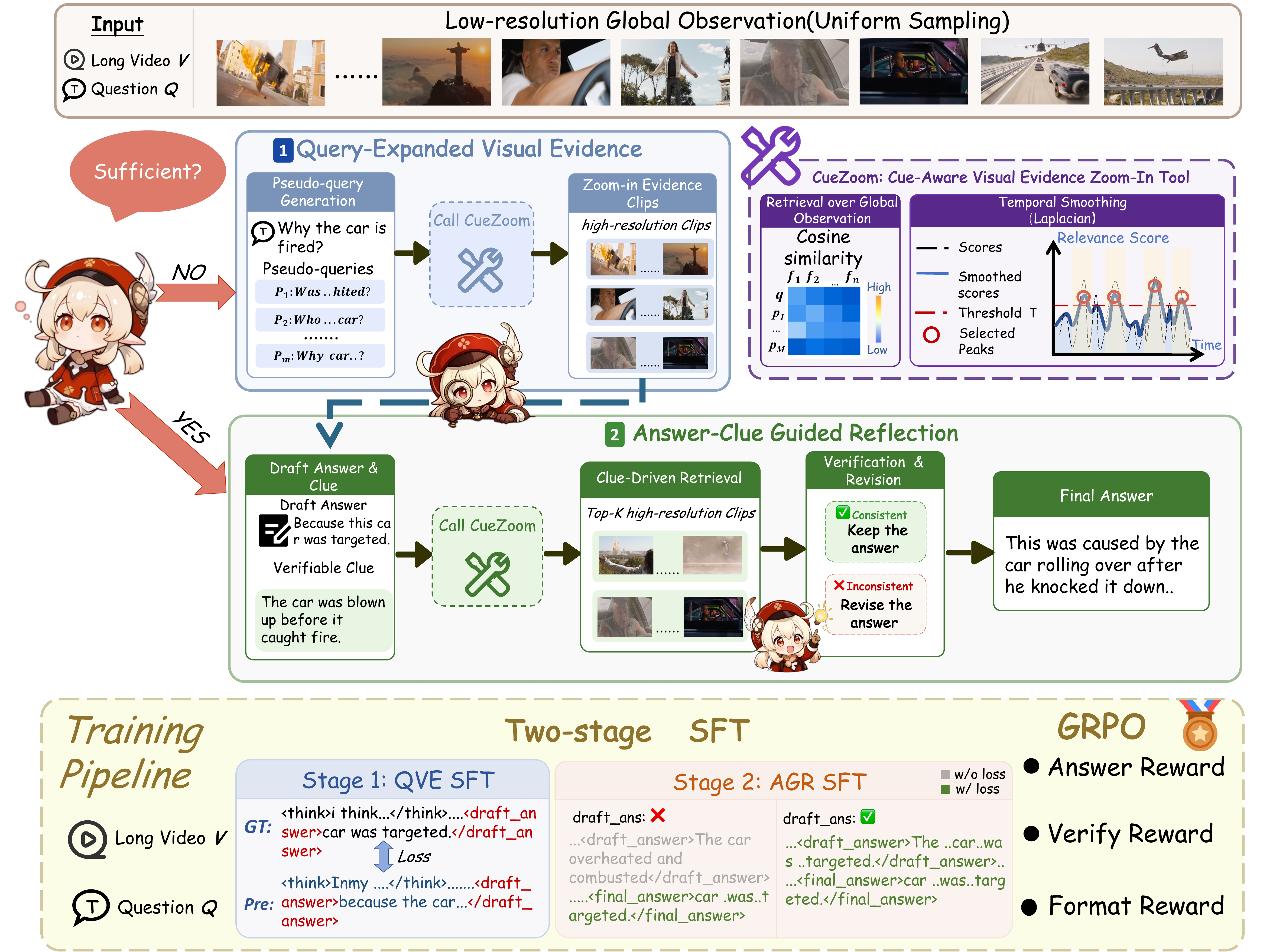}
    \caption{
\textbf{The overview of the proposed CoVER framework.}
CoVER first uses pseudo-queries to retrieve task-relevant zoom-in clips for evidence-aware draft generation, and then uses an answer-specific clue to retrieve verification evidence for final-answer refinement.
The model is trained with staged supervised fine-tuning and evidence-guided reinforcement learning using answer, verification, and format rewards.
}
    \label{fig:main}
\end{figure*}
\section{Method}
\subsection{Overview}
We propose CoVER, a \textbf{Co}mprehensive \textbf{V}isual \textbf{E}vidence and \textbf{R}eflection framework for long-video understanding. CoVER enables the model to acquire comprehensive visual evidence before answer generation and further reflect on draft answers with answer-specific visual clues. Instead of relying only on a low-resolution global observation, CoVER combines query-expanded visual evidence acquisition with answer-clue guided visual reflection, allowing the model to reason over both holistic video context and fine-grained visual evidence.

As illustrated in Fig.~\ref{fig:main}, the model first builds a low-resolution global observation of the input video $V$ to capture holistic context with reduced computation. Based on the question $q$ and the global observation, it decides whether to invoke the visual zoom-in tool for fine-grained evidence acquisition. When additional evidence is needed, the model generates pseudo-queries $P=\{p_m\}_{m=1}^{M}$ from the question and global observation. These pseudo-queries act as evidence-seeking queries that describe potentially missing visual facts. Together with the original question, they guide the zoom-in tool to retrieve evidence clips from the full video. The retrieved query-expanded evidence complements the global observation and supports draft answer generation. After producing a draft answer, the model derives an answer-specific visual clue and invokes the zoom-in tool again to retrieve verification evidence. It then checks whether the draft answer is visually supported and either preserves or revises it. Thus, CoVER forms a closed-loop reasoning process that acquires missing evidence, verifies draft predictions, and refines answers through explicit visual feedback. Overall, CoVER consists of two components: \emph{Query-Expanded Visual Evidence (QVE)} for draft answer generation, and \emph{Answer-Clue-Guided Reflection (AGR)} for answer verification and refinement.

\subsection{CueZoom: Cue-Aware Visual Evidence Zoom-In Tool}

To support evidence acquisition and verification in CoVER, we introduce \textbf{CueZoom}, a cue-aware visual evidence zoom-in tool. CueZoom retrieves visual evidence clips according to textual cues, including the original question, pseudo-queries, and answer-specific clues. Given textual cues $\mathcal{U}=\{u_j\}_{j=1}^{M}$ and sampled video frames $\{f_i\}_{i=1}^{N}$, CueZoom computes the relevance score $s_i$ of each frame $f_i$ by averaging its similarity to all cues:
\begin{equation}
s_i = \frac{1}{M}\sum_{j=1}^{M}
\mathrm{sim}\big(\phi_t(u_j), \phi_v(f_i)\big),
\end{equation}
where $\phi_t(\cdot)$ and $\phi_v(\cdot)$ are the text and visual encoders, respectively, and $\mathrm{sim}(\cdot,\cdot)$ denotes cosine similarity.
Since frame-level relevance scores can be noisy and may contain isolated responses, we apply Laplacian Temporal Smoothing (LTS) to obtain more stable visual evidence:
\begin{equation}
\tilde{s}_i = \sum_{j=-r}^{r} \kappa_j s_{i+j},
\quad
\kappa_j \propto \exp(-|j|/h),
\end{equation}
where $\tilde{s}_i$ denotes the smoothed relevance score of frame $f_i$, $\kappa_j$ is the smoothing weight for its neighboring offset $j$, $r$ is the smoothing radius, and $h$ controls the decay rate of the smoothing kernel. Local normalization is applied near video boundaries. We then select frames whose smoothed scores exceed an adaptive threshold $\tau=\mu(\tilde{s})+\alpha\sigma(\tilde{s})$, where $\mu(\tilde{s})$ and $\sigma(\tilde{s})$ are the mean and standard deviation of the smoothed relevance sequence, and $\alpha$ controls the selectivity of evidence selection. Finally, neighboring selected frames are merged into coherent visual evidence clips.

\subsection{Query-Expanded Visual Evidence}

Given a long video $V$ and a question $q$, the low-resolution global observation $V_{\mathrm{global}}$ provides overall semantic context, but may miss sparse or fine-grained evidence. To improve evidence coverage before answer generation, we introduce \emph{Query-Expanded Visual Evidence (QVE)}. Specifically, the model generates a set of pseudo-queries $P=\{p_d\}_{d=1}^{D}$ from $q$ and $V_{\mathrm{global}}$, where $D\leq3$ in our implementation. These pseudo-queries act as evidence-seeking queries that describe question-relevant visual facts to be further verified. 
We combine the original question and pseudo-queries into expanded textual cues: $\mathcal{U}_{\mathrm{qe}}=\{q\}\cup P$. CueZoom uses these cues to locate and retrieve relevant video clips from the full video. The retrieved clips $\mathcal{C}_{\mathrm{qe}}$ serve as high-resolution query-expanded visual evidence for draft answer generation.

\subsection{Answer-Clue Guided Reflection}

Although query-expanded visual evidence improves draft answer generation, the draft answer may still be insufficiently verified by visual evidence. To address this, we introduce \emph{Answer-Clue-Guided Reflection (AGR)}. Given a draft answer $\hat{a}$, the model derives an answer-specific clue $c$, which describes the visual condition expected to hold if $\hat{a}$ is correct. 
We use $c$ as the verification cue: $\mathcal{U}_{\mathrm{ver}}=\{c\}.$
CueZoom then retrieves answer-relevant evidence clips based on this cue. Unlike pseudo-query-based retrieval, which expands question-relevant evidence coverage, clue-guided retrieval focuses on evidence that directly supports or contradicts the draft answer. This enables the model to reflect on $\hat{a}$ using explicitly retrieved visual evidence rather than treating it as final.

\subsection{Data Construction and Two-Stage SFT}
\begin{figure}[t]
    \centering
    \includegraphics[width=\columnwidth]{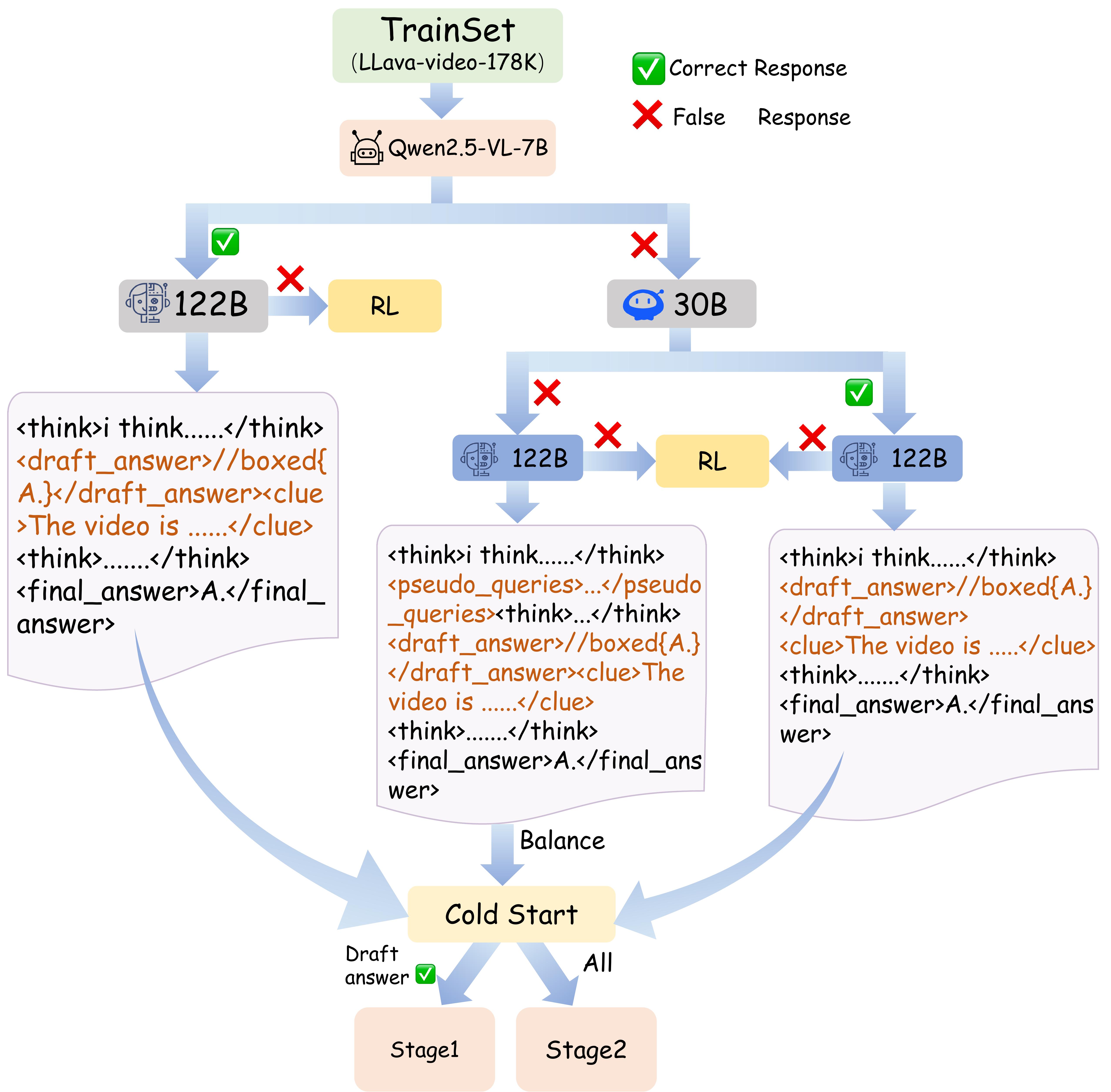}
    \caption{
\textbf{Training data construction pipeline.}
Samples are partitioned by difficulty and annotated into structured trajectories. Correct draft-answer trajectories are used for Stage-1 QVE SFT, while all annotated trajectories support Stage-2 AGR SFT; incorrectly answered samples are further used for RL optimization.
}
    \label{fig:data_pipeline}
\end{figure}

\begin{figure}[t]
    \centering
    \includegraphics[width=\linewidth]{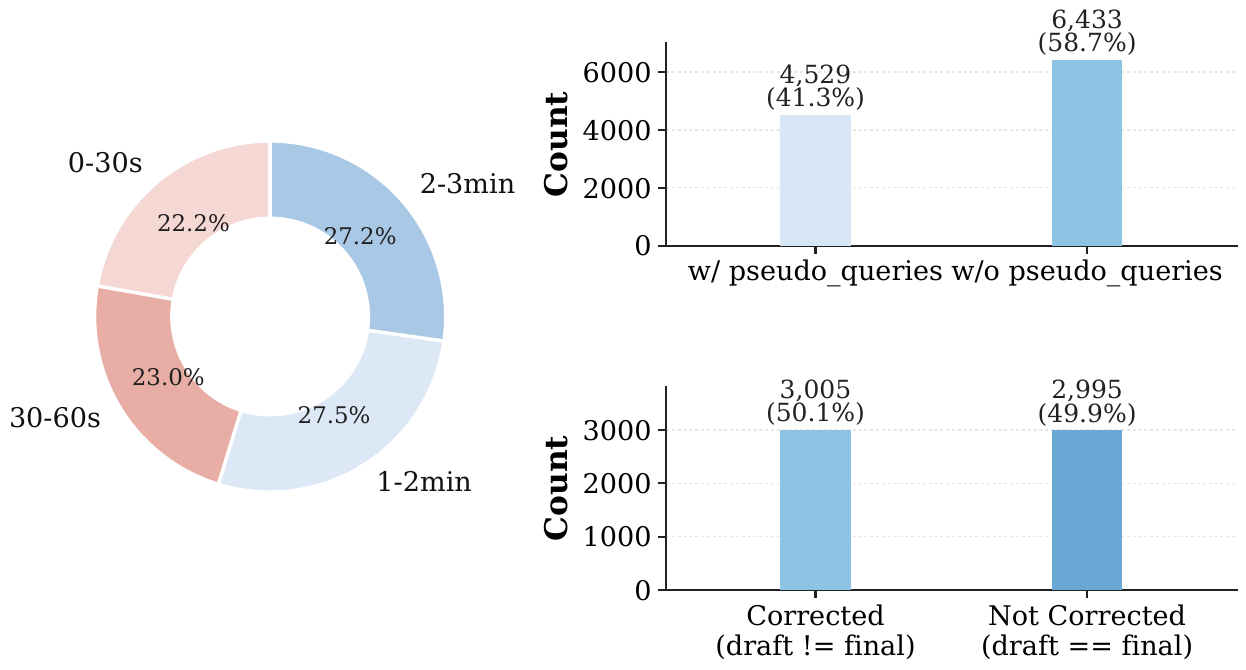}
    \caption{
\textbf{Statistics of the constructed training data.}
The dataset covers diverse video durations, both pseudo-query and direct-answer trajectories, and balanced corrected/no-corrected draft-answer cases.
}
    \label{fig:data_dis}
\end{figure}

\noindent
\textbf{Data Construction.} To provide a cold start for RL, we construct 11K high-quality SFT samples from LLaVA-Video-178K~\cite{zhang2024llava} through the automated pipeline shown in Fig.~\ref{fig:data_pipeline}. The pipeline consists of three steps: (1) Data Partitioning. We first evaluate each sample with Qwen2.5-VL-7B. Samples correctly answered by Qwen2.5-VL-7B are treated as easier cases, while failed samples are further processed by Qwen3-VL-30B-A3B to identify tool-dependent or harder cases. This partition allows the dataset to cover both direct-answer and evidence-seeking reasoning patterns. (2) CoT Annotation Generation. Qwen3.5-122B-A10B is used to generate structured reasoning traces, including draft answers, pseudo-queries when additional evidence is needed, visual clues, retrieved evidence, and final verified answers. These traces provide supervision for both pseudo-query guided evidence acquisition and clue-guided answer verification. (3) Data Balancing and Cleaning. We balance different trace types, upsample rare answer-revision cases where retrieved evidence corrects an incorrect draft answer, and apply rule-based filtering to remove malformed or inconsistent samples. The detailed distribution of the dataset is 
show in Fig.~\ref{fig:data_dis}.

\noindent
\textbf{Stage-1: QVE SFT.}
This stage aims to train the model to generate reliable draft answers from available visual evidence. Cold-start trajectories with correct draft answers are used as Stage-1 data. Given the global observation and question, the model learns an adaptive tool-use policy: it invokes the tool with pseudo-queries to retrieve additional visual evidence when needed, and bypasses the tool to directly produce a draft answer when the available evidence is sufficient. In both cases, the model also generates an answer-specific visual clue for subsequent verification. This stage initializes evidence-aware draft prediction while avoiding supervision from erroneous intermediate answers.

\noindent

\noindent
\textbf{Stage-2: AGR SFT.}
This stage trains the model to verify draft answers with answer-clue guided evidence. Given the question, global observation, draft answer, answer clue, and verification clips, the model learns to assess whether the draft answer is visually supported and decide whether the final answer should be revised.
To avoid optimizing toward unreliable intermediate drafts, we mask the loss on incorrect draft answers while keeping supervision for verification reasoning and final-answer generation:

\begin{equation}
\begin{aligned}
\mathcal{L}_{\mathrm{AGR}}
&=
-\sum_{t=1}^{T}
m_t \log p_{\theta}(y_t \mid x, y_{<t}), \\
m_t
&=
\begin{cases}
0, & \neg z_{\hat{a}} \land (t < t_{\mathrm{da}}), \\
1, & \text{otherwise},
\end{cases}
\end{aligned}
\end{equation}
where $T$ is the sequence length, $y_t$ is the target token at position $t$, $x$ is the input context, and $y_{<t}$ denotes preceding target tokens. The binary mask $m_t$ controls whether token $y_t$ contributes to the loss. $z_{\hat{a}}$ indicates whether the draft answer $\hat{a}$ is correct, and $t_{\mathrm{da}}$ denotes the position of the closing \texttt{</draft\_answer>} tag. When the draft answer is incorrect, tokens before this tag are masked out.

\subsection{GRPO with Evidence-Guided Rewards}
We adopt Group Relative Policy Optimization (GRPO)~\cite{guo2025deepseek} to encourage dynamic and autonomous tool use during reinforcement learning. Given a generated reasoning trajectory $\tau$, GRPO jointly optimizes final-answer correctness, answer-level verification, and format consistency through a composite reward:
\begin{equation}
R(\tau)
=
\lambda_{\mathrm{ans}} R_{\mathrm{ans}}
+
\lambda_{\mathrm{ver}} R_{\mathrm{ver}}
+
\lambda_{\mathrm{fmt}} R_{\mathrm{fmt}},
\label{eq:dr_grpo_reward}
\end{equation}
where $\lambda_{\mathrm{ans}}$, $\lambda_{\mathrm{ver}}$, and $\lambda_{\mathrm{fmt}}$ are reward weights. We set them to $0.6$, $0.25$, and $0.15$, respectively. 

\noindent

\textbf{Answer Reward}~\cite{yu2026dapo}.
The answer reward \(R_{\mathrm{ans}}\) serves as the primary task-level signal for encouraging correct final predictions by comparing the final answer with the ground-truth.

\noindent
\textbf{Verification Reward.}
The verification reward measures whether the transition from the draft answer to the final answer is consistent with evidence-grounded verification. 
Given the extracted draft answer $d$, final answer $f$, and label answer $y$:
\begin{equation}
R_{\mathrm{ver}}=
\begin{cases}
1, & f=y,\\
-0.5, & d=y \ \text{and}\ f\neq y,\\
-0.5, & d=\emptyset,\\
-1, & f=\emptyset,\\
0, & \text{otherwise}.
\end{cases}
\label{eq:verification_reward}
\end{equation}
This reward assigns a positive score when the final answer is correct, regardless of whether the model preserves a correct draft or revises an incorrect one. 
It penalizes cases where a correct draft is revised into an incorrect final answer, thereby discouraging unsupported answer changes. 
Missing draft or final answers are also penalized to encourage complete verification trajectories.

\noindent\textbf{Format Reward}~\cite{zhang2025r1}.
The format reward \(R_{\mathrm{fmt}}\) encourages the model to follow the predefined response format.

\section{Experiments}
\subsection{Experimental Setup}

\noindent
\textbf{Implementations.} We instantiate our framework with Qwen2.5-VL-7B~\cite{bai2025qwen2} as the base model. For supervised fine-tuning, we adopt two stages: Stage-1 is trained for 3 epochs to establish evidence-aware reasoning, and Stage-2 for 1 epoch to enhance final-answer verification and correction. For reinforcement learning, we further train with GRPO~\cite{guo2025deepseek} on 4K high-quality reasoning samples using the proposed evidence-guided rewards. All experiments are conducted on 8 NVIDIA A800 GPUs, with more details reported in the Appendix.

\noindent
\textbf{Benchmarks for Evaluation.}
We conduct a comprehensive experimental analysis to assess the video reasoning capability of our method.
For a broad evaluation, we adopt representative video understanding benchmarks, including MLVU~\cite{zhou2025mlvu}, Video-MME~\cite{fu2025video}, LongVideoBench~\cite{wu2024longvideobench}, and LVBench~\cite{wang2025lvbench}.
MLVU and Video-MME evaluate general video understanding and reasoning.
LongVideoBench and LVBench focus on long-video comprehension, evidence aggregation, and temporal-semantic reasoning.
\begin{table*}[!t]
\centering
\caption{
\textbf{Comparison with SOTA video-LLMs on long-video benchmarks.}
We compare CoVER with proprietary, open-source, agentic, and reasoning-oriented models on MLVU, VideoMME, LongVideoBench, and LVBench. * denotes our reproduced result. 
$\Delta$ reports the performance improvement of CoVER over baseline.
768L+32H denotes 768 low-resolution and 32 high-resolution frames.
}
\label{tab:main_results}
\vspace{-0.5em}
\setlength{\tabcolsep}{2.2pt}
\renewcommand{\arraystretch}{0.92}
\resizebox{0.85\textwidth}{!}{
\begin{tabular}{@{}lcccccccc@{}}
\toprule
Models & Size & \#Frames & Context & MLVU & \multicolumn{2}{c}{VideoMME} & LongVideoBench & LVBench \\
\cmidrule(lr){6-7}
       &      &          &         & 3$\sim$120 min & Overall & Long & 0$\sim$60 min & 4101 sec \\
\midrule

\rowcolor{groupblue}
\multicolumn{9}{@{}l@{}}{\hspace{0.6em}\textit{Proprietary Models}} \\
GPT-4V & -- & 1fps & -- & -- & 60.7 & 56.9 & -- & -- \\
GPT-4o & -- & 1fps & -- & 66.2 & 77.2 & 72.1 & 66.7 & 34.7 \\
\midrule

\rowcolor{groupblue}
\multicolumn{9}{@{}l@{}}{\hspace{0.6em}\textit{Open-Source Video MLLMs}} \\
Video-LLaVA~\cite{lin2024video} & 7B & 8 & -- & 47.3 & 40.4 & 38.1 & 39.1 & -- \\
LLaMA-VID~\cite{li2024llama} & 7B & 1fps & -- & 33.2 & -- & -- & -- & 23.9 \\
ShareGPT4Video~\cite{chen2024sharegpt4video} & 8B & 16 & -- & 46.4 & 43.6 & 37.9 & 39.7 & -- \\
LLaVA-NeXT-Video~\cite{zhang2024llavanextvideo} & 7B & 32 & -- & -- & 46.5 & -- & 43.5 & -- \\
VideoLLaMA2~\cite{cheng2024videollama} & 7B & 32 & -- & 48.5 & 46.6 & 43.8 & -- & -- \\
LongVA~\cite{zhang2024long} & 7B & 128 & -- & 56.3 & 54.3 & 47.6 & -- & -- \\
VideoChat2~\cite{li2024mvbench} & 7B & 16 & -- & 47.9 & 54.6 & 39.2 & -- & -- \\
LLaVA-OneVision~\cite{li2024llava} & 7B & 32 & -- & 64.7 & 58.2 & 46.7 & -- & -- \\
Vamba~\cite{ren2025vamba} & 10B & 1024 & -- & 65.9 & 57.8 & -- & 55.9 & 42.1 \\
VideoChat-T~\cite{zeng2025timesuite} & 7B & 12 & -- & -- & 46.3 & 41.9 & -- & -- \\
Video-XL~\cite{shu2025video} & 7B & 256 & -- & 64.9 & 55.5 & -- & 50.7 & -- \\
Video-XL-Pro~\cite{liu2025video} & 7B & 240 & -- & 70.6 & 60.0 & -- & 56.7 & -- \\
LongVILA~\cite{chen2025longvila} & 7B & 256 & -- & -- & 60.1 & 53.0 & 57.1 & -- \\
LongVU~\cite{shen2024longvu} & 7B & 1fps & -- & 65.4 & 60.6 & 59.5 & -- & -- \\
Hour-LLaVA~\cite{lin2026unleashing}& 7B & 1fps & -- & -- & 63.6 & 55.0 & 60.4 & 45.6 \\
LongVITA-128K~\cite{shen2025long} & 14B & 256 & -- & -- & 66.4 & 58.8 & 60.9 & -- \\
VILAMP~\cite{cheng2025scaling} & 7B & 1fps & -- & 72.6 & 67.5 & 57.8 & 61.2 & 45.2 \\
VideoChat-Flash~\cite{li2024videochat} & 7B & 512 & -- & 74.7 & 65.3 & 55.4 & -- & 48.2 \\
\midrule

\rowcolor{groupblue}
\multicolumn{9}{@{}l@{}}{\hspace{0.6em}\textit{Open-Source Agent Video MLLMs}} \\
VideoMind~\cite{bhatnagar2026videomind} & 7B & -- & -- & 64.4 & 58.2 & 49.2 & 56.3 & 40.8 \\
Video-RAG~\cite{luo2026video} & 7B & -- & -- & 72.4 & 62.1 & 59.8 & 58.7 & -- \\
\midrule

\rowcolor{groupblue}
\multicolumn{9}{@{}l@{}}{\hspace{0.6em}\textit{Open-Source Reasoning Video MLLMs}} \\
Video-MTR~\cite{xie2025video} & 7B & 32 & 4k & 48.4 & 59.0 & 51.0 & -- & -- \\
Video-R1~\cite{feng2026video} & 7B & 32 & 4k & -- & 59.3 & -- & -- & -- \\
VITAL~\cite{zhang2025thinking} & 7B & 1024 & 32k & -- & 64.1 & 54.0 & -- & -- \\
LongVILA-R1~\cite{chen2026scaling} & 7B & 512 & 132k & -- & 65.1 & 55.2 & 58.0 & -- \\
Conan~\cite{ouyang2025conan} & 7B &32& -- & 63.4 & 60.5 & -- & 56.6 & 39.2 \\
\midrule

\rowcolor{groupblue}
\multicolumn{9}{@{}l@{}}{\hspace{0.6em}\textit{Ours}} \\
Qwen2.5-VL*~\cite{bai2025qwen2} & 7B & 128 & 16k & 66.4 & 63.1 & 53.6 & 56.0 & 42.0 \\
\textbf{CoVER} & \textbf{7B} & 768L + 32H & \textbf{16k} & \textbf{70.3} & \textbf{63.8} & \textbf{54.7} & \textbf{58.3} & \textbf{46.6} \\
$\Delta$ & -- & -- & -- & +3.9 & +0.7 & +1.1 & +2.3 & +4.6 \\
\bottomrule
\end{tabular}
}
\vspace{-0.8em}
\end{table*}

\subsection{Main Results}

Table~\ref{tab:main_results} provides a comprehensive comparison of CoVER with representative video MLLMs across four benchmarks, including open-source, agentic, and reasoning-oriented models such as Video-MTR~\cite{xie2025video} and Conan~\cite{ouyang2025conan}. For a fair comparison, Qwen2.5-VL-7B is evaluated with its default inference setting and a comparable visual-token budget using 128 uniformly sampled frames at approximately \(448\times448\) resolution. CoVER consistently outperforms this baseline, achieving gains of +3.9\% on MLVU, +0.7\% on VideoMME, +1.1\% on VideoMME-Long, +2.3\% on LongVideoBench, and +4.6\% on LVBench. CoVER also outperforms Video-MTR and 
Conan on their reported metrics. These results demonstrate the effectiveness of query-expanded visual evidence  and answer-clue guided reflection for evidence-aware video reasoning.

CoVER maintains strong general video understanding on MLVU and VideoMME and achieves larger gains on LongVideoBench and LVBench, where sparse evidence requires visual verification. Its competitive performance against 7B-scale reasoning models shows that explicit evidence acquisition and clue-guided verification improve long-video reasoning without increasing model scale.

\begin{table}[t]
\centering
\small
\setlength{\tabcolsep}{2.5pt}
\renewcommand{\arraystretch}{1.05}
\caption{
\textbf{Ablation on training stages.} LongVB denotes LongVideoBench.
}
\label{tab:training_stage_ablation}
\begin{tabular}{lccccc}
\toprule
Model & MLVU & VideoMME & LongVB & LVBench \\
\midrule
\textbf{CoVER} & \textbf{70.3} & \textbf{63.8} & \textbf{58.3} & \textbf{46.6} \\
w/o Stage-1 SFT & 67.5 & 62.9 & 57.0 & 45.8  \\
w/o Stage-2 SFT & 66.9 & 63.0 & 56.2 & 42.2 \\
w/o RL & 63.3 & 61.5 & 56.4 & 42.6 \\
w/o SFT + RL & 43.4 & 50.3 & 40.8 & 31.0 \\
\bottomrule
\end{tabular}
\end{table}
\subsection{Ablation Study}
\noindent
\textbf{Ablation Study on Training Stages.}
Table~\ref{tab:training_stage_ablation} evaluates the contribution of each training stage. CoVER achieves the best performance across all benchmarks, reaching 70.3 on MLVU, 63.8 on VideoMME, 58.3 on LongVideoBench, and 46.6 on LVBench. Removing either Stage-1 SFT or Stage-2 SFT  degrades performance, showing that both stages are important for evidence-grounded reasoning. In particular, removing Stage-2 SFT causes a 4.4\% drop on LVBench, indicating the importance of supervised verification for long-video understanding. Removing RL leads to a larger drop than removing either SFT stage on MLVU and LVBench, highlighting its role in further optimizing evidence-grounded reasoning after supervised training. The variant without SFT or RL performs worst, showing all training stages are important.

\begin{table}[t]
\centering
\small
\setlength{\tabcolsep}{2.9pt}
\caption{
\textbf{Ablation study on CoVER components.}
LongVB denotes LongVideoBench.
}
\label{tab:reflection_ablation}
\begin{tabular}{lcccc}
\toprule
Model & MLVU & VideoMME& LongVB & LVBench \\
\midrule
w/o QVE & 66.4 & 63.1 &  56.0 & 44.0 \\
w/o AGR & 67.4 & 63.3  & 57.1 &45.3\\
\textbf{CoVER} & \textbf{70.3} & \textbf{63.8}  & \textbf{58.3} & \textbf{46.6} \\
\bottomrule
\end{tabular}
\end{table}

\begin{table}[!h]
\centering
\footnotesize
\setlength{\tabcolsep}{2.6pt}
\renewcommand{\arraystretch}{0.82}
\caption{
\textbf{CueZoom hyperparameter sensitivity on LVBench.}
Each hyperparameter is varied independently, with the default setting 
\((r=4, h=1, \alpha=0.5)\) highlighted in gray.
}
\vspace{-6pt}
\label{tab:sensitivity-r-h-alpha}
\begin{tabular}{@{}c@{\hspace{3pt}}|@{\hspace{3pt}}ccc@{\hspace{3pt}}|@{\hspace{3pt}}ccc@{\hspace{3pt}}|@{\hspace{3pt}}ccc@{}}
\toprule
\textbf{Param.} 
& \multicolumn{3}{c|}{$r$} 
& \multicolumn{3}{c|}{$h$} 
& \multicolumn{3}{c}{$\alpha$} \\
\midrule
\textbf{Val.}
& 0 & \cellcolor{gray!15}\textbf{4} & 8
& 0.5 & \cellcolor{gray!15}\textbf{1} & 2
& 0.25 & \cellcolor{gray!15}\textbf{0.5} & 1 \\
\textbf{Score}
& 43.8 & \cellcolor{gray!15}\textbf{46.6} & 45.7
& 46.0 & \cellcolor{gray!15}\textbf{46.6} & 46.0
& 45.6 & \cellcolor{gray!15}\textbf{46.6} & 45.0 \\
\bottomrule
\end{tabular}
\vspace{-6pt}
\end{table}

\begin{figure*}[t]
    \centering
    \includegraphics[width=0.9\linewidth]{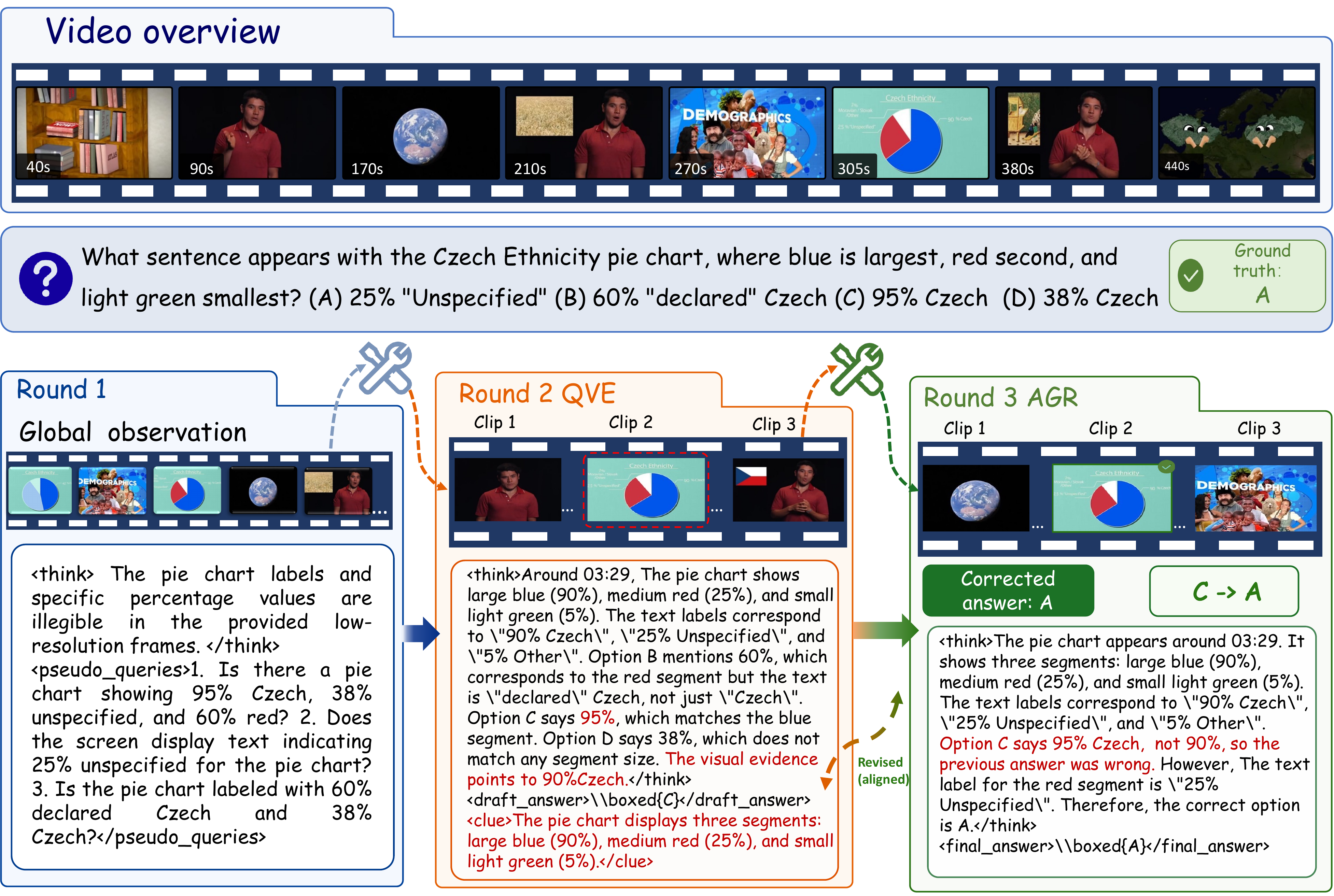}
    \caption{
\textbf{Case study of CoVER's reasoning process.}
This example illustrates CoVER’s two-step evidence process: pseudo-query based zoom-in for fine-grained evidence retrieval, followed by answer-specific verification to revise an unsupported draft into the correct answer.
} 
    \label{fig:case_study}
\end{figure*}

\noindent\textbf{Ablation Study on CoVER Components.}
Table~\ref{tab:reflection_ablation} ablates CoVER's two core components: Query-Expanded Visual Evidence (QVE) and Answer-Clue Guided Reflection (AGR). 
Removing QVE or AGR consistently degrades performance across all benchmarks, showing that both components contribute to CoVER's final performance. 
Compared with w/o QVE, CoVER improves by +3.9 on MLVU, +0.7 on VideoMME, +2.3 on LongVideoBench, and +2.6 on LVBench. 
Compared with w/o AGR, CoVER further improves by +2.9, +0.5, +1.2, and +1.3, respectively.

\subsection{Hyperparameter Sensitivity.}
We analyze the sensitivity of CueZoom Tool's three hyperparameters: the smoothing radius $r$, Laplacian bandwidth $h$, and adaptive threshold coefficient $\alpha$. As shown in Table~\ref{tab:sensitivity-r-h-alpha}, the default setting ($r=4$, $h=1$, $\alpha=0.5$) achieves the best performance. Removing temporal smoothing ($r=0$) clearly hurts performance, while overly large radius values also lead to degradation. This indicates that moderate smoothing helps suppress noisy peaks without blurring fine-grained evidence. Results for $h$ and $\alpha$ further highlight the importance of proper temporal aggregation and thresholding.

\subsection{Case Study}
Fig.~\ref{fig:case_study} shows a representative case of CoVER's evidence-grounded reasoning. 
When the low-resolution global observation is insufficient, pseudo-queries help retrieve fine-grained zoom-in evidence that improves evidence coverage. 
The answer-specific clue then guides verification toward evidence directly related to the draft answer, enabling CoVER to identify unsupported predictions and revise them. 
This case highlights the complementary roles of QVE and AGR in finding missing evidence and verifying answer grounding. 
Additional qualitative results, including more case studies and representative failure cases, are provided in the appendix.

\section{Conclusion}
We present \textbf{CoVER}, a visual evidence and reflection framework for long-video understanding. CoVER improves Video-LLM reasoning from two aspects: it gathers more complete visual evidence through query-expanded evidence acquisition, and it verifies answers through answer-clue guided visual reflection. By integrating these two modules, CoVER shifts long-video understanding from open-loop answer generation to evidence-centric and visually verifiable reasoning. Experiments show that CoVER improves over the baseline, demonstrating the importance of comprehensive evidence acquisition and visual-feedback reflection for reliable long-video reasoning.

\section{Limitations}
Although CoVER improves evidence-grounded long-video reasoning, it still depends on the quality of generated evidence-seeking queries and answer-specific clues. Ambiguous or biased queries may retrieve incomplete or misleading evidence. In addition, answer-clue-guided reflection is conditioned on the initial answer, so incorrect initial predictions may lead to biased verification clues. The zoom-in based evidence retrieval process also introduces extra computation compared with direct uniform sampling. Finally, CoVER may be less effective for questions that require holistic video understanding rather than localized visual evidence. Future work can improve query robustness, tool invocation, and retrieval efficiency.
\bibliography{custom}

\end{document}